\documentclass[11pt,letterpaper]{article}
\usepackage{naaclhlt2016}
\usepackage{times}
\usepackage{latexsym}

\naaclfinalcopy 

\title{Duluth at SemEval--2016 Task 14 : \\ Extending Gloss Overlaps to Enrich Semantic Taxonomies}
\author{Ted Pedersen\\
    Department of Computer Science\\
    University of Minnesota\\
    Duluth, MN, 55812 USA\\
    {\tt tpederse@d.umn.edu}}

\date{}

\begin{document}

\maketitle

\begin{abstract}
This paper describes the Duluth systems that participated in Task 14 of 
SemEval 2016, Semantic Taxonomy Enrichment. There were three related
systems in the formal evaluation which are discussed here, 
along with numerous post--evaluation runs. All of
these systems identified synonyms between WordNet and other dictionaries
by measuring the gloss overlaps between them. These systems perform
better than the random baseline and one post--evaluation variation was within
a respectable margin of the median result attained by all participating 
systems. 
\end{abstract}

\section{Introduction}

The goal of Task 14 in SemEval--2016 was to enrich a semantic taxonomy with 
new word senses. In particular, this task sought to augment 
WordNet\footnote{http://wordnet.princeton.edu}
with senses that are present in another dictionary. Task 14 drew glosses of
words not found in WordNet from a variety of sources that will collectively
be referred to as OtherDict in this paper.

The method the Duluth systems took was based on scoring overlaps between
WordNet and OtherDict glosses. A OtherDict sense was assigned to the
WordNet sense with the highest overlapping score. Overlaps are defined
to be exact matches between words and phrases in the glosses. Each
overlap is assigned a score that is the square of the number of words
in the overlap, and then all the overlaps between a pair of glosses is
summed to provide the final score. 

The premise of relying on overlaps is that senses that are defined
using many of the same words are certainly related, and indeed if they
are defined using the same words they are likely synonyms. This is a well 
established and reliable intuition that goes back at least to \cite{Lesk86}. 
Closely related words that may not be synonyms (such as hyponyms or 
hypernyms) will use many of the same words in their definitions, but then
have specific differentia that distinguishes among them. 

Task 14 asked participants to distinguish between merging a OtherDict sense into an
existing WordNet synset (i.e, a synonym) or attaching it as a hyponym 
(i.e., a more specific example) of a WordNet sense. However, our systems only
merge OtherDict senses into WordNet synsets. It seems clear though that 
this merge versus attach problem can be tackled by setting some kind of 
threshold for the amount of overlap, where more significant degrees of overlap 
should result in a synonym merge whereas less overlap could indicate a hyponym attach. 
As yet we have not been successful in determining a reliable method for finding
such a threshold. As a result we simply assumed every OtherDict sense would attach
to its closest (most overlapping) WordNet sense. 

Each Duluth system carried out the same pre-processing on both the WordNet
and OtherDict glosses. In addition, the WordNet glosses were extended using
additional information from WordNet such as the glosses of its hypernyms, hyponyms, 
derived forms, and meronyms. This follows naturally from the structure of  
WordNet and the intuitions that underlie the Extended Gloss Overlap measure
 \cite{BanerjeeP03B}, which is implemented in WordNet::Similarity \cite{PedersenPM04a} and 
UMLS::Similarity \cite{McInnesPP09}. Unfortunately there was not time to expand
the OtherDict glosses in similar ways, although this is at least a possibility since
some other dictionaries (such as Wiktionary) provide hyponyms and hypernyms, 
among other relations. 

Task 14 allowed two kinds of systems to participate : {\em resource--aware}
that only used dictionary content, and {\em constrained} 
that used other resources beyond dictionaries. The Duluth systems 
are considered resource--aware since they only use information from WordNet and
OtherDict.

\section{Systems}

All the Duluth systems start by pre--processing both WordNet and OtherDict glosses.
This consists of removing any character that is not alphanumeric, and then 
converting all remaining characters to lower case. Compounds known to WordNet are 
identified in the WordNet glosses, but not as it turns out in the 
OtherDict glosses. This very likely reduced the number of overlaps we found
since a WordNet compound such as {\em light\_year} will not match {\em light year}, which
is the form that would occur in OtherDict. 

In the following sections we describe the Duluth systems. Since the only distinction between
them is how they reconstruct WordNet glosses we will provide a running example to illustrate 
each system. We will use the noun feline\#n\#1 for this purpose. The original WordNet 
entry for this sense (prior to pre--processing) is shown here : 

\begin{itemize}
\item feline\#n\#1 -- (any of various lithe-bodied roundheaded fissiped mammals, many with retractile claws)
\item hypernym : carnivore -- (a terrestrial or aquatic flesh-eating mammal; ``terrestrial carnivores have four or five clawed digits on each limb")
\item hyponym1 : cat, true cat -- (feline mammal usually having thick soft fur and no ability to roar: domestic cats; wildcats)
\item hyponym2 : big cat, cat -- (any of several large cats typically able to roar and living in the wild)
\item meronym : feline -- (a clawed foot of an animal especially a quadruped)
\item derived : feline\#a\#1 -- (of or relating to cats; ``feline fur")
\end{itemize}

The Duluth systems build upon each other to some extent, so they are presented in 
an order that more easily illustrates those connections rather than their numeric
order (which has no particular significance).

\subsection{Duluth2}


Duluth2 is the most basic of the Duluth systems. Each WordNet sense is 
represented by its gloss where all stop words and single character 
words have been removed. The stoplist comes from the Ngram Statistics 
Package \cite{BanerjeeP03} and includes 392 
words\footnote{http://cpansearch.perl.org/src/TPEDERSE/Text-NSP-1.31/bin/utils/stoplist-nsp.regex}.

To continue our example, the first noun sense of feline is represented by Duluth2 as shown 
below - it has simply gone through pre--processing and then had stop words removed.

\begin{itemize}
\item feline\#n\#1  :  lithe bodied roundheaded fissiped mammals 
retractile claws
\end{itemize}

This system represents a baseline for the overlap measures, since we are 
comparing the original WordNet and OtherDict glosses after having done
minimal pre--processing. 

\subsection{Duluth1}


Duluth1 is a natural extension of Duluth2, where each WordNet gloss is expanded
by concatenating to it (in the following order) : the glosses of the hypernyms of 
the sense\footnote{In general each sense has only one hypernym, although this is not
always true in WordNet.},
the glosses of all the hyponyms of the sense, the glosses of any derived form 
of the sense, and the glosses of all the meronyms of the sense. 
This significantly expands
the size of each WordNet gloss, to the point where our initial attempts to simply 
use these extended glosses in matching took too much time to finish during the available
window of time for the evaluation. We were also concerned about the significant disparities
in size among WordNet glosses, and of course with the unexpanded OtherDict glosses. 

As a result we decided to shorten the 
WordNet glosses in Duluth1 by removing any word that is made up of four characters or less 
(rather than using a stoplist), 
and then only taking the first nine words in the expanded gloss. 
For many 
words this means that just the original gloss and the gloss of the hypernym
and perhaps part of the gloss of a hyponym would be included. The OtherDict glosses
were processed in a similar fashion, where any word with 4 or fewer characters was removed.

Our running example is shown below. Note that this is quite similar to Duluth2, except 
that {\em various} is included below (but was excluded in Duluth2 since it is
in the stoplist), and that {\em terrestrial} is included (since it is
the first word in the gloss of the hyponym). 

\begin{itemize}
\item feline\#n\#1  :  various lithe bodied roundheaded fissiped mammals 
retractile claws terrestrial  
\end{itemize}

\subsection{Duluth4}


Note that Duluth4 was not included in our official evaluation. Rather this was run after 
the evaluation period, and it proved to be our most accurate result. Duluth4 is similar to 
Duluth1 in that it expands the WordNet glosses, but only does so with the glosses of its hypernyms
and hyponyms (and does not include the derivational forms or meronyms, as Duluth1 does). Stop
words are removed using the same list as Duluth2. 

We can see in our running example that this provides a larger gloss, but it is not as large
as what Duluth1 provided (before pruning it back to just the first nine words). 

\begin{itemize}
\item feline\#n\#1  :   lithe bodied roundheaded fissiped mammals retractile claws terrestrial aquatic flesh 
eating mammal terrestrial carnivores four five clawed digits limb feline mammal having thick soft fur ability 
roar domestic cats wildcats several large cats typically able roar living wild 
\end{itemize}

\subsection{Duluth3}


There are many minor variations between words in WordNet and OtherDict glosses, and it
is difficult to normalize the glosses in order to eliminate them. Instead, we decided to
have one system that relied on character tri-grams, since that could allow for matches
between portions of words, rather than requiring the exact matches that all the other 
Duluth systems insist upon. 

Duluth3 expands each WordNet gloss with the gloss of its hypernyms and its hyponyms (like Duluth4,
although stop words are not eliminated in Duluth3). Then all spaces are removed from each expanded
gloss, and it is broken into three character ngrams. Glosses were limited to 250 of these trigrams,
mainly so that they could finish running in the available time during the evaluation. 

If one studies the running example carefully you can reconstruct the gloss, which is similar to
Duluth4 except that it includes stop words.

\begin{itemize}
\item feline\#n\#1  :   any  ofv  ari  ous  lit  heb  odi  edr  oun  dhe  ade  dfi  ssi  ped  mam  mal  sma  nyw  ith  ret  rac  til  ecl  aws  ate  rre  str  ial  ora  qua  tic  fle  she  ati  ngm  amm  alt  err  est  ria  lca  rni  vor  esh  ave  fou  ror  fiv  ecl  awe  ddi  git  son  eac  hli  mbf  eli  nem  amm  alu  sua  lly  hav  ing  thi  cks  oft  fur  and  noa  bil  ity  tor  oar  dom  est  icc  ats  wil  dca  tsa  nyo  fse  ver  all  arg  eca  tst  ypi  cal  lya  ble  tor  oar  and  liv  ing  int  hew  ild  
\end{itemize}

\section{Results and Discussion}

\begin{table}
\begin{center}
\label{table:results}
\begin{tabular}{l|rrrr} 
     	& Wu \& 	& Lemma		& 	 &  \\ 
	& Palmer 	& Match		& Recall & F1 \\ \hline
First Word  & .5140 & .4150 & 1.00 &  .6790 \\
Median      &       &       &      &   .5900 \\
Duluth4(*)     & .3810 & .0550 & 1.00 &  .5518 \\
Duluth2     & .3471 & .0433 & 1.00 &  .5153 \\
Duluth3     & .3452 & .0167 & 1.00 &  .5132 \\
Duluth1     & .3312 & .0233 & 1.00 &  .4976 \\
Random      & .2269 & .0000 & 1.00 &  .3699 \\
\end{tabular}
\caption{Task 14 results, (*) indicates post--evaluation run.}
\end{center} 
\end{table}

Despite seemingly significant differences in how the WordNet glosses were
expanded, Table 1 reveals that there were only minor
differences in results among Duluth1, Duluth2, and Duluth3. 
This seems to support the conclusion that gloss overlaps 
provide a reliable and robust starting point for this problem. Indeed, the simplest of 
our approaches (Duluth2, which used WordNet glosses minus stop words) 
was slightly more effective than two more elaborate variations (Duluth1 and Duluth3). 
Recall that Duluth1 normalized gloss lengths by taking only the first nine
words in the expanded glosses, and that Duluth3 attempted a kind of poor 
man's stemming by using character trigrams. 

Duluth4 is an obvious extension of Duluth2, in that it expands glosses
with their hypernym and hyponyms. This results in a modest but significant
improvement on the systems submitted for the formal evaluation, and 
suggests that focusing on expanded gloss content in relatively 
straightforward ways can pay dividends. This system at least begins to approach
the median score attained by the systems participating in the task. 
That said, even this result considerably lags the baseline First Word \cite{JurgensP15} 
provided by the organizers. 

Table 2 shows the results of additional experiments were carried out with Duluth1.
Here the gloss sizes were varied from 1 to 100, where 9 is the value used during 
the formal evaluation. Note that the size of the OtherDict glosses was always 5,533 
(tokens) and that the recall attained was always 1.00.

\begin{table}
\begin{center}
\label{table:results2}
\begin{tabular}{l|rrrrrr} 
gloss  	& token & Wu \& 	& Lemma		&  \\ 
size	& count & Palmer 	& Match		 & F1 \\ \hline
1 & 195,242 & 0.3033 & 0.0050 & 0.4654  \\
5 & 973,463 & 0.3336 & 0.0183 & 0.5003  \\
9 & 1,701,884  & 0.3312 & 0.0233 & 0.4976 \\
10 & 1,866,198 & 0.3278 & 0.0200 & 0.4937 \\
20 & 3,023,538  & 0.3296 & 0.0200 & 0.4958  \\
30 & 3,575,586  & 0.3476 & 0.0300 & 0.5159  \\
40 & 3,871,786  & 0.3471 & 0.0383 & 0.5153  \\
50 & 4,059,495  & 0.3444 & 0.0400 & 0.5123  \\
100 & 4,466,581  & 0.3511 & 0.0400 & 0.5197 \\
\end{tabular}
\caption{Duluth 1 post--evaluation variations.}
\end{center} 
\end{table}

Table 2 shows that generally speaking increasing the number
of words in the WordNet glosses results in a small but steady improvement
in performance. However, it is important to keep in mind that as the
WordNet glosses are growing, the OtherDict glosses are fixed at the same
size. This seems to make it clear that the most important avenue moving forward
is to expand the OtherDict glosses with additional content, comparable to
that with which WordNet is expanded. 

\section{System Implementation Details}

The Duluth systems are implemented using UMLS::Similarity, which provides measures
of semantic relatedness and similarity for the Unified Medical Language System. In addition,
it also allows a user supplied dictionary to be automatically
included and utilized by the Lesk measure. Thus, the Duluth systems created 
dictionaries from OtherDict and WordNet suitable for use by the Lesk measure in 
UMLS::Similarity (via the --dict option). 
WordNet was accessed using the Perl module 
WordNet::QueryData\footnote{http://search.cpan.org/WordNet-QueryData}.

\end{document}